\documentclass[sigplan,nonacm]{acmart}

\AtBeginDocument{%
  \providecommand\BibTeX{{%
    \normalfont B\kern-0.5em{\scshape i\kern-0.25em b}\kern-0.8em\TeX}}}

\usepackage[switch]{lineno}

\usepackage{booktabs} %
\usepackage{tabularx}
\usepackage{adjustbox}
\usepackage{framed}
\usepackage{enumitem}

\usepackage{ragged2e}
\newcolumntype{P}[1]{>{\RaggedRight\hspace{0pt}}m{#1}}
\usepackage{color, colortbl}
\definecolor{Gray}{gray}{0.94}
\usepackage{pifont}
\usepackage[capitalize]{cleveref}
\usepackage{tcolorbox}
\newcommand{\reforms}{\textsc{reforms}}

\makeatletter
\def\@copyrightspace{\relax}
\let\@authorsaddresses\@empty
\makeatother
\setcopyright{none}
\renewcommand\footnotetextcopyrightpermission[1]{} %
\author{Sayash Kapoor$^1$
\hspace{1em}  Emily Cantrell
\hspace{1em}  Kenny Peng
\hspace{1em} Thanh Hien Pham \\
\hspace{1em} Christopher A. Bail 
\hspace{1em}  Odd Erik Gundersen
\hspace{1em}  Jake M. Hofman
\hspace{1em}  Jessica Hullman \\
\hspace{1em}  Michael A. Lones 
\hspace{1em}  Momin M. Malik
\hspace{1em}  Priyanka Nanayakkara
\hspace{1em}  Russell A. Poldrack \\
\hspace{1em}  Inioluwa Deborah Raji
\hspace{1em}  Michael Roberts 
\hspace{1em}  Matthew J. Salganik
\hspace{1em}  Marta Serra-Garcia \\
\hspace{1em}  Brandon M. Stewart
\hspace{1em}  Gilles Vandewiele 
\hspace{1em}  Arvind Narayanan
\\
\bigskip
\textbf{September 19, 2023}
\\
\bigskip
}

\date{}

\begin{document}
\sloppy
\title{\LARGE \reforms{}: Reporting Standards for Machine Learning Based Science}
\renewcommand{\shortauthors}{Kapoor et al.}
\renewcommand{\shorttitle}{\reforms{}: Reporting Standards for Machine Learning Based Science}

\begin{abstract}
    Machine learning (ML) methods are proliferating in scientific research. However, the adoption of these methods has been accompanied by failures of validity, reproducibility, and generalizability. 
    These failures can hinder scientific progress, lead to false consensus around invalid claims, and undermine the credibility of ML-based science.
    ML methods are often applied and fail in similar ways across disciplines.
    Motivated by this observation, our goal is to provide clear reporting standards for ML-based science. 
    Drawing from an extensive review of past literature, we present the \reforms{} checklist (\textbf{Re}porting Standards \textbf{For} \textbf{M}achine Learning Based \textbf{S}cience). It consists of 32 questions and a paired set of guidelines. 
    \reforms{} was developed based on a consensus of 19 researchers across computer science, data science, mathematics, social sciences, and biomedical sciences.
    \reforms{} can serve as a resource for researchers when designing and implementing a study, for referees when reviewing papers, and for journals when enforcing standards for transparency and reproducibility.
\end{abstract}

\maketitle
\footnotetext[1]{Corresponding author: sayashk@princeton.edu. The latest version of the paper and the appendices are available at \href{https://reforms.cs.princeton.edu/}{https://reforms.cs.princeton.edu}}

\section*{Introduction}

ML methods are being widely adopted for scientific research \cite{athey_machine_2019,schrider_supervised_2018,valletta_applications_2017,iniesta_machine_2016,tonidandel_big_2018, yarkoni_choosing_2017, grimmer_machine_2021, leist_mapping_2022, wiemken_machine_2020, varian_big_2014, mullainathan_machine_2017}.
Compared to older statistical methods, they offer increased predictive accuracy \cite{athey_machine_2019}, the ability to process large amounts of data \cite{breiman_statistical_2001}, and the ability to use different types of data for scientific research, such as text, images, and video~\cite{grimmer_machine_2021}. 
However, the rapid uptake of ML methods has been accompanied by concerns of validity, reproducibility, and generalizability \cite{hullman_worst_2022, beam_challenges_2020, roberts_common_2021, bouthillier_unreproducible_2019, mcdermott_reproducibility_2021, gundersen_state_2018, varoquaux_machine_2022}.
There are several reasons for concern. Performance evaluation is notoriously tricky in ML \cite{schmidt_descending_2021, bouthillier_accounting_2020, demasi_meaningless_2017, sculley_winners_2018}. 
ML code tends to be complex and as yet lacks standardization \cite{liu_successes_2019,dittmer_navigating_2023}, leading to a lack of computational reproducibility~\cite{gundersen_machine_2022}.
Subtle pitfalls arise from the differences between explanatory and predictive modeling \cite{hofman_prediction_2017}. 
The hype and over-optimism about commercial AI may spill over into scientific research \cite{banja_ai_2020}. 
In addition, publication biases that have led to past reproducibility crises \cite{johnson_reproducibility_2017} are also present in ML research \cite{bell_perspectives_2021, pineau_improving_2022}.
If left unchecked, these flaws can lead to a feedback loop of overoptimism since non-replicable findings are cited more than replicable ones \cite{serra-garcia_nonreplicable_2021}.
There is an urgent need to systematically address errors in ML-based science rather than finding errors in individual studies after publication. 

In this paper, we focus on a specific subset of ML applications: ML-based science. We use this term to refer to research that makes a scientific claim using the performance of an ML model as evidence. For example, Salganik et al. \cite{salganik2020measuring} use ML methods to investigate the predictability of life outcomes. This contrasts with ML methods research, which involves improving widely applicable ML methods instead of making scientific claims using ML methods. In the next section, we clarify the distinctions between ML methods research and ML-based science and outline the scope of the paper in greater detail. Box 1 summarizes this discussion.

One promising way to detect and prevent errors in scientific research is by improving reporting standards \cite{mongan_checklist_2020, bossuyt_stard_2015, white_strengthening_2015}. Clear expectations for using ML methods can allow researchers and referees to spot errors early. 
Despite the use of ML methods across disciplines, there are no widely applicable best practices for reporting the design, implementation, and evaluation of ML-based science.
This leads to different, and often no fixed reporting standards in each field adopting ML methods. As a result, common failure modes in using ML methods recur across disciplines \cite{simmonds_how_2022, kapoor_leakage_2022}.

In this paper, we introduce a set of reporting standards for ML-based science, with the goal of preventing known but common errors that occur when ML is used in scientific research.
To that end, we review the literature on best practices and common errors in ML-based science. We introduce the \reforms{} checklist and guidelines for reporting ML-based science, which consists of 32 items across 8 modules (Table 2). We accompany \reforms{} with a detailed set of guidelines to set expectations for each item (Appendix B).

Checklists have been adopted in many scientific fields \cite{mongan_checklist_2020, bossuyt_stard_2015, white_strengthening_2015, collins_transparent_2015}, and they have been impactful in improving reporting practices \cite{plint_does_2006, han_checklist_2017}. In 2014, the U.S. National Institutes of Health (NIH) created principles to improve reproducibility and rigor, endorsed by several journals. One item was the creation of reporting standards and checklists for journals \cite{noauthor_principles_2015}. The EQUATOR network collects reporting guidelines for health research and includes over 500 checklists \cite{noauthor_reporting_nodate}. Several checklists have been proposed in ML methods research \cite{pineau_improving_2022, rogers_just_2021, gundersen_reproducible_2018}.

\reforms{} differs from this large body of past work in crucial ways.
    First, past checklists for scientific research are field or method-specific. For example, the \textit{CLAIM} checklist~\cite{mongan_checklist_2020} provides best practices for reporting research on AI in medical imaging. As a result, many items in the checklist do not apply to other scientific fields adopting ML methods. Since ML methods are being rapidly adopted across fields, and research that uses ML methods suffers from similar errors, we aimed to make the \reforms{} checklist field-agnostic. We selected items that broadly apply to fields that use ML methods. 
    Second, past checklists for ML methods research focus on common errors in developing ML methods. But these errors differ from the ones that arise in scientific research (See Box 1 for the distinction between ML methods research and ML-based science). For instance, Pineau et al.'s checklist \cite{pineau_improving_2022} does not include questions about the distribution about which the claims are made, since ML methods research often focuses on improving a model's performance on a benchmark dataset~\cite{donoho_50_2017}. In contrast, clearly specifying the distribution of interest is core to a scientific claim. 
Still, past work in both scientific research and ML methods research has helped inform our checklist.

\enlargethispage{\baselineskip}

We present findings from an extensive review of past research on best practices and common shortcomings relevant to each of the 8 checklist modules. Alongside the checklist, we release guidelines paired with each item in the checklist. Similar guidelines have been included in past efforts at establishing reporting standards. For instance, the STROBE-RDS statement \cite{white_strengthening_2015} included an ``explanation and elaboration'' document to clarify expectations for each item in the reporting standards. Our guidelines aim to increase the usability of the \reforms{} checklist and to provide pointers to best practices for ML-based science. 

The \reforms{} checklist can help address common failure modes that lead to errors in ML-based science. 
To guard against corners being cut due to time and publication pressures, the standards provide a set of clear expectations.
To aid researchers new to ML-based science, the guidelines for each item identify resources and relevant past literature. While no single document would be enough to familiarize researchers with all the nuances of ML-based science, our hope is that the guidelines can be one useful pedagogical resource. 
Finally, there are many steps involved in successfully reporting ML-based science. It is hard to keep all of these items in mind when writing up a study. \reforms{} provides all 32 items in one document to prevent omission errors.

\section*{The scope of our claims: ML-based science}

We define ML-based science as scientific research that uses the performance of a machine learning model as evidence for a scientific claim. This includes, but is not limited to, making predictions, conducting measurements, or performing other tasks that contribute to the body of scientific knowledge.
This definition has two parts: first, the research should be geared toward answering a scientific question of interest. This means that other types of ML research and applications do not fit under the umbrella of ML-based science:

\begin{figure}[!t]
\begin{tcolorbox}[colback=gray!20,colframe=gray!40,rounded corners]
\section*{Box 1: What is ML-based science?}

ML-based science refers to scientific research that uses machine learning (ML) models to contribute to scientific knowledge. This includes making predictions, conducting measurements, or performing other tasks that help answer scientific questions. 

\medskip
ML-based science could use a variety of ML techniques such as supervised learning, unsupervised learning, reinforcement learning, and deep learning. The research should be geared toward answering a scientific question of interest.

\medskip
\textbf{Exclusions.} Not all ML research and applications qualify. For example, ML methods research and predictive analytics fall outside the scope. Similarly, quantitative research not using ML methods, such as explanatory modeling and simulations, do not qualify as ML-based science.

\medskip
\textbf{Applicability.} Our checklist for ML-based science will be more useful for some types of research than others. For instance, research on the use of ML for predictions may benefit more than research on using ML for search tasks within vast and complex spaces.

\end{tcolorbox}
\end{figure}

\begin{itemize}[leftmargin=*]
\item \textbf{ML methods research.} Research focusing on developing and refining machine learning methods, such as a typical NeurIPS paper, does not constitute ML-based science. While such work does contribute new ML methods, the main objective is not to solve specific scientific problems of interest about generalizable populations. (Still, elements of the \reforms{} checklist could be valuable to ML methods research. This is particularly the case when these newly developed methods are evaluated on benchmarks that directly influence their application in scientific contexts despite not being representative of real problems.)
\item \textbf{Predictive analytics. }Many real-world applications of ML models emphasize predictive accuracy but are not conducted to gain scientific insights. For example, social media platforms use ML to predict if a user will click on ads millions of times a day. 
There many differences between predictive analytics and ML-based science. In predictive analytics, the population or distribution of interest may not be clearly defined. The relationships found using ML models only need to hold within a company or organization, and need not generalize. Relative accuracy is more important than absolute accuracy numbers since the decision to deploy a model has often already been made---so the only question being answered by the modeling activity is \textit{which} of a given set of models should be deployed in production. And verifying if the predictions later come true is often easy. This feedback is a better test for whether an application works as intended (compared to using a checklist).
\end{itemize}

Second, the research should utilize ML methods. By this, we refer to a variety of techniques including, but not limited to, supervised learning, unsupervised learning, reinforcement learning, and deep learning. This even includes foundation models~\cite{bommasani2022opportunities}, which are a type of ML model, although our focus in this paper is more general.~\footnote{There are several challenges with using proprietary foundation models for scientific research. For example, they are non-deterministic and can change without adequate notice~\cite{kapoor_openais_2023}. This could lead to hard-to-resolve shortcomings in computational reproducibility. Foundation models are one example of a grey area in the definition of ML-based science. In such cases, we leave it up to the authors and referees to decide if the \reforms{} checklist is useful.} 
Consequently, other types of quantitative research not employing these methods do not fall under the umbrella of ML-based science. 

\begin{itemize}[leftmargin=*]
\item \textbf{Explanatory modeling. }Such research utilizes traditional statistical methods for improving our understanding of real-world phenomena rather than predicting outcomes~\cite{hofman_integrating_2021}. This is an example of quantitative scientific research that does not use ML methods. While explanatory modeling suffers from many shortcomings~\cite{hullman_worst_2022}, errors in this domain are subtly different compared to ML-based science, which means much of our checklist does not apply to explanatory modeling research.
\item \textbf{Simulation. } 
Similarly, physics-based models and simulation methods are sometimes evaluated iteratively by output and by their fidelity with existing theory~\cite{winsberg2010, pfeffer2017}, and do not involve machine learning or fitting models to data---except potentially as a tool to understand and summarize the simulation results. As a result, many items in the \reforms{} checklist do not apply. Note that some algorithms used in machine learning, like Markov Chain Monte Carlo, are examples of numerical simulations; but these are used only to fit models to data, which is different simulating an underlying phenomenon of interest.

\end{itemize}

We call research at the intersection of ML methods and quantitative science ``ML-based science''. Salganik et al.'s work on the Fragile Families Challenge~\cite{salganik2020measuring} is an example of this category. They used machine learning to predict children's life outcomes and answer scientific questions about the predictability of outcomes studied by sociologists. We discuss many other examples in our review.

Note that our checklist will likely be more helpful for some types of ML-based science than others. For instance, it is likely to be more useful for predictive modeling compared to research that uses machine learning methods for search tasks in vast and complex spaces, such as the search for new materials or new phases of matter~\cite{carrasquilla_machine_2017}. In cases where verifying the result of an ML-based experiment is easy (for instance, verifying the properties of a new drug in a lab), our checklist might be less useful than such verification, though it could help ensure the validity of the experiments prior to verification.

\begin{tcolorbox}[float,floatplacement=!t,colback=gray!20,colframe=gray!40,rounded corners]
\section*{Box 2: Goals of the \reforms{} checklist}

\medskip

\medskip
\textbf{Goal 1: Establish the scientific claim and its relation to the ML task.} A key feature of our checklist, distinguishing it from those used in ML methods research, is its focus on using ML to support scientific claims. In such research, it is necessary to establish the intended scientific claim as well as how the performance of the ML model supports that claim. For example, it is necessary to state the scientific claim's population of interest and then justify why the dataset used in the ML task represents this population.	This should be compared to ML methods research, where the performance of a model is often itself the claim.

\medskip
\textbf{Goal 2: Ensure that the ML task is executed correctly and that the performance is reported in sufficient detail.} To establish that the performance of the specified ML model supports the intended scientific claim, it is necessary to ensure that the performance of the model is calculated correctly. There are many ways in which the performance of a model can be misleading. For example, a common error in ML research is evaluating a model on data it was trained on, resulting in overly optimistic results. Additionally, carefully reporting uncertainty is necessary to interpret model performance correctly.	

\medskip
\textbf{Goal 3: Enable an independent scientist to verify results. }Finally, our checklist is designed to help ensure that all resources and descriptions needed for verifying a study are provided alongside the paper. Thus, our checklist helps ensure that independent researchers can understand and evaluate a given study.

\medskip
These goals are not intended to be disjoint and often support one another. They can help orient the reader when navigating our checklist, and they reveal how our checklist is tailored to ML-based science. 

\end{tcolorbox}

\section*{Methods}

To develop the \reforms{} checklist, we started with a focus on steps in a canonical ML pipeline, drew from previous checklists used in other domains, and went through a consensus process with all authors, involving multiple rounds of feedback and a virtual discussion. Table 1 lists the modules in our checklist and the corresponding stages in the ML pipeline. For each module, we focus on three goals: (1) establishing the scientific claim and its relationship to the ML modeling process; (2) providing an overview of the best practices and common shortcomings in building ML models correctly; and (3) enabling the verification of the results by an independent researcher. In other words, we aim to decrease the likelihood of errors of interpretation (goal 1) or execution (goal 2), and to make it easier for independent researchers to spot errors (goal 3). Box 2 outlines these goals in more detail.

\begin{table}[!b]
    \centering
    \renewcommand{\arraystretch}{1.2} 
    \begin{adjustbox}{width=\columnwidth,center}
    \begin{tabular}{p{2.5cm}|p{7cm}}
        \toprule
        \textbf{Stage of scientific study} & \textbf{Section of the checklist}
        \\\midrule
        \textbf{Study design} & Study goals (Module 1) \newline Computational reproducibility (Module 2)
        \\ \midrule \rowcolor{Gray}
        \textbf{Data collection and preparation} & Data quality (Module 3) \newline Data preprocessing (Module 4)
        \\ \midrule 
        \textbf{Modeling} & Modeling decisions (Module 5)
        \\ \midrule \rowcolor{Gray}
        \textbf{Evaluation} & Data leakage (Module 6) \newline Metrics and uncertainty quantification (Module 7)
        \\\midrule 
        \textbf{Scope and limitations} & Generalizability and limitations (Module 8)
        \\ \bottomrule
    \end{tabular}
    \end{adjustbox}
    \caption{Stages of ML-based science and corresponding checklist sections.}
    \label{table:study-stages}
    \vspace{-0.3cm}
\end{table}

To build on previous efforts at improving the reporting quality of research, we used three past checklists to ensure our coverage of important items in reporting an ML model.
Pineau et al. \cite{pineau_improving_2022} provided the checklist used alongside papers submitted to NeurIPS 2020, a prominent ML methods conference. 
Collins et al. \cite{collins_transparent_2015} provided the \textit{TRIPOD} checklist for prediction models in health research.
Mongan et al. \cite{mongan_checklist_2020} introduced the \textit{CLAIM} checklist for AI models in clinical imaging.
We chose these checklists because they covered diverse modeling approaches and were applicable in different settings (ML methods research, models for individual diagnosis and prognosis, and ML for medical imaging, respectively).

\begin{figure*}[!t]
\begin{tcolorbox}[colback=gray!20,colframe=gray!40,rounded corners]
\section*{Box 3: How authors, referees, and journals can use \reforms{}}

\medskip
Our reporting standards can help improve the quality of ML-based science in multiple ways. 

\begin{itemize}[leftmargin=*]

\item \textbf{Authors} can self-regulate by using the \reforms{} to identify errors and preemptively address concerns about using ML methods in their paper \cite{han_checklist_2017}. This can also help increase the credibility of their paper, especially in fields that are newly adopting ML methods. We expect that \reforms{} will be useful to authors throughout the study---during conceptualization, implementation, and communication of the results (See Table 1 for a list of checklist modules corresponding to these goals). The checklist can be included as part of the supplementary materials released alongside a paper. The guidelines can help authors learn how to correctly apply the \reforms{} checklist in their own work and introduce them to underlying theories of evidence.

\item \textbf{Referees} can use \reforms{} to determine whether a study they are reviewing falls short. If they have concerns about a study, they can ask researchers to include the filled-out checklist in a revised version. For example, Roberts et al. \cite{roberts_common_2021} use the \textit{CLAIM} checklist \cite{mongan_checklist_2020} to filter papers for a systematic review based on compliance with the checklist.

\item \textbf{Journals} can require authors to submit a checklist along with their papers to set reporting standards for ML-based science. Similar checklists are in place in a number of journals \cite{nature_reporting_nodate, science_science_nodate}; however, they are usually used for specific disciplines rather than for methods that are prevalent across disciplines \cite{macleod_mdar_2021}. Since ML-based science is proliferating across disciplines, \reforms{} offers a method-specific (rather than discipline-specific) intervention.

\end{itemize}

Note that the \reforms{} checklist is additive to field-specific norms. It is not a replacement for existing requirements within fields, such as pre-registration, ethics reviews, or using discipline-specific checklists for other parts of the research process. It might seem burdensome to ask researchers to adhere to another set of standards, but our work was born out of painful necessity: studies across fields have repeatedly found reproducibility errors in ML-based science \cite{kapoor_leakage_2022}, and in the absence of a systematic intervention, this is likely to worsen. 

\end{tcolorbox}
\end{figure*}

\textbf{Consensus process for developing the \reforms{} checklist.} Once we had an initial set of items, authors met virtually for a discussion. One of the main outcomes of the discussion was the need for a paired set of guidelines alongside each item to clarify how these items should be reported, which we discuss in more detail below. Then, the authors collaboratively edited the checklist to choose commonly applicable items across disciplines. We paid close attention to usability: to decrease the time and cognitive load that using the checklist would entail, we removed items that were too specific and would apply to a small subset of ML-based science. Finally, the authors independently flagged unclear items to improve the quality of the reporting standards.

A recurring theme in our conversation was making the reporting standards easy to use. To that end, we developed a set of accompanying guidelines to help researchers understand the motivation for each section and clarify what is expected for each item in \reforms{}. The guidelines (Appendix B) are based on our review of past literature for items in the checklist (this review is presented next). We include references to key prior work to help onboard researchers new to using ML methods. This includes a mix of peer-reviewed scientific research that details best practices, as well as resources from the ML methods community that outline common shortcomings and ways to address them. Crucially, we do not take prescriptive stances on matters of ongoing methodological debate. Instead, we present best practices to minimize and detect known types of errors in ML-based science.

\begin{table*}[!t]
  \centering
  \begin{adjustbox}{width=\textwidth,center}
\begin{tabular}{p{3cm}|p{16cm}}
\toprule
\textbf{Section} & \textbf{Item}
\\\midrule
\textbf{Study goals} & 1a. Population or distribution about which the scientific claim is made
\\ \rowcolor{Gray}
& 1b. Motivation for choosing this population or distribution \textit{(1a.)}
\\
& 1c. Motivation for the use of ML methods in the study
\\ \midrule
\rowcolor{Gray}
\textbf{Computational} & 2a. Dataset used for training and evaluating the model along with link or DOI to uniquely identify the dataset 
\\
\textbf{reproducibility}& 2b. Code used to train and evaluate the model and produce the results reported in the paper along with link or DOI to uniquely identify the version of the code used 
\\ \rowcolor{Gray}
& 2c. Description of the computing infrastructure used
\\
& 2d. README file which contains instructions for generating the results using the provided dataset and code
\\ \rowcolor{Gray}
& 2e. Reproduction script to produce all results reported in the paper  
\\ \midrule 
\textbf{Data quality} & 3a. Source(s) of data, separately for the training and evaluation datasets (if applicable), along with the time when the dataset(s) are collected, the source and process of ground-truth annotations, and other data documentation 
\\ \rowcolor{Gray}
& 3b. Distribution or set from which the dataset is sampled (i.e., the sampling frame)
\\
& 3c. Justification for why the dataset is useful for the modeling task at hand 
\\ \rowcolor{Gray}
& 3d. The outcome variable of the model, along with descriptive statistics (split by class for a categorical outcome variable) and its definition. 
\\
& 3e. Sample size and outcome frequencies
\\ \rowcolor{Gray}
& 3f. Percentage of missing data, split by class for a categorical outcome variable
\\
& 3g. Justification for why the distribution or set from which the dataset is drawn \textit{(3b.)} is representative of the one about which the scientific claim is being made \textit{(1a.)}
\\ \midrule \rowcolor{Gray}
\textbf{Data preprocessing} & 4a. Identification of whether any samples are excluded with a rationale for why they are excluded
\\
& 4b. How impossible or corrupt samples are dealt with
\\ \rowcolor{Gray}
& 4c. All transformations of the dataset from its raw form \textit{(3a.)} to the form used in the model, for instance, treatment of missing data and normalization. Preferably through a flow chart
\\ \midrule \rowcolor{Gray}
\textbf{Modeling} & 5a. Detailed descriptions of all models trained
\\
& 5b. Justification for the choice of model types implemented
\\ \rowcolor{Gray}
& 5c. Method for evaluating the model(s) reported in the paper, including details of train-test splits or cross-validation folds
\\
& 5d. Method for selecting the model(s) reported in the paper
\\ \rowcolor{Gray}
& 5e. For the model(s) reported in the paper, specify details about the hyperparameter tuning
\\
& 5f. Justification that model comparisons are against appropriate baselines
\\ \midrule \rowcolor{Gray}
\textbf{Data leakage} & 6a. Justification that pre-processing (Section 4) and modeling (Section 5) steps only use information from the training dataset (and not the test dataset)
\\
& 6b. Methods to address dependencies or duplicates between the training and test datasets (e.g. different samples from the same patients are kept in the same dataset partition)
\\ \rowcolor{Gray}
& 6c. Justification that each feature or input used in the model is legitimate for the task at hand and does not lead to leakage
\\ \midrule \rowcolor{Gray}
\textbf{Metrics and uncertainty} & 7a. All metrics used to assess and compare model performance (e.g., accuracy, AUROC etc.). Justify that the metric used to select the final model is suitable for the task
\\
& 7b. Uncertainty estimates (e.g., confidence intervals, standard deviations), details of how these are calculated 
\\ \rowcolor{Gray}
& 7c. Justification for the choice of statistical tests (if used) and a check for the assumptions of the statistical test
\\ \midrule \rowcolor{Gray}
\textbf{Generalizability} & 8a. Evidence of external validity
\\
\textbf{and limitations}& 8b. Contexts in which the authors do not expect the study’s findings to hold
\\
\bottomrule
\end{tabular}
  \end{adjustbox}
\caption{
The \reforms{} checklist for ML-based science. Some items are shortened for brevity; see Appendix A for a complete template. Alongside each item, authors should report the section or page number where the item is reported. Some items in the \reforms{} checklist could be hard to report for specific studies. Instead of requiring strict adherence for each item, authors and referees should decide which items are relevant for a study and how details can be reported better. To that end, we hope that the checklist can offer a useful starting point for authors and referees working on ML-based science. 
}
  \label{table:checklist}
\end{table*}

\textbf{Organization of the paper.} In the remainder of the paper, we present the \reforms{} reporting standards. They comprise eight modules based on the stages of an ML-based science study (see \cref{table:study-stages}). For each module, we motivate why items in this module are important to address in ML-based science. For each item, we include expectations about what it means to address the item sufficiently. In our review, we draw from past literature on best practices and common errors in ML-based science. 
We also occasionally draw on literature about science with traditional statistical methods, as best practices and shortcomings are shared in many aspects of ML-based science and other quantitative science. 
In Appendix A, we provide a template for the \reforms{} checklist. In Appendix B, we distill the guidelines for filling out the checklist as a standalone document. In Appendix C, we present a table with additional details on the references used in our review. 

\section*{Module 1: Study goals}

This section focuses on stating a study’s goals. This is motivated by recent research which shows that reporting study goals in adequate depth and clarity is not trivial or common \cite{lundberg_what_2021}. Studies that appear to ask the same research question may actually have subtle differences in their questions which lead to substantially different findings ~\cite{Hofman2017}.

\textbf{1a) Population or distribution about which the scientific claim is made.} The "population of interest" is the group to which the researchers intend the findings of the study to generalize. This is typically broader than the sampling frame and sample, which are discussed in Module 3. Defining the population of interest is important because it shapes the article's conclusions, places boundaries on those conclusions, and provides the basis for metrics of significance and uncertainty that derive from the concept of sampling from a population \cite{Wilkinson1999StatisticalMI, lundberg_what_2021, Casteel2021}. Unfortunately, research articles do not always clearly state their population of interest \cite{Tooth2005, lundberg_what_2021, Simons2017}.

\textbf{1b) Motivation for choosing this population or distribution.} The choice of a particular population of interest may be motivated by pure scientific interest or by a need for applied knowledge. Explaining the motivation for studying the population of interest helps the reader understand the importance of the study and contextualize its results.  

We acknowledge that the motivation for choosing a particular population of interest may arise from what data is available. The development of the motivation depends on whether researchers followed a deductive, inductive, or iterative approach \cite{Grimmer_Roberts_Stewart_2022}. In a deductive approach, researchers begin with a theory and a population of interest, then select or create a dataset based on the data’s ability to test that theory for that population. In an inductive approach, researchers begin with a dataset, then determine what research questions and populations of interest that dataset can address. In an iterative approach, researchers iterate between data collection, data analysis, and theory until they develop a hypothesis, and then collect additional data to test that hypothesis \cite{Grimmer_Roberts_Stewart_2022}. The deductive approach is the most widely accepted approach in the scientific community, but Grimmer et al. argue that there is also great value in inductive and iterative approaches, and that it is important to communicate which approach was used \cite{Grimmer_Roberts_Stewart_2022}.

\textbf{1c) Motivation for the use of ML methods in the study.} The research questions asked in ML-based studies often differ from those asked with traditional statistical methods. Breiman famously argued that there are "two cultures" in statistical modeling: the "data modeling" culture and the "algorithmic modeling" culture. In the "data modeling" culture, which is aligned with traditional statistical methods, researchers’ focus is on estimating the parameters of a function that is meant to meaningfully represent the process by which input data produces output data in the world. For example, researchers might ask, "what is the relationship between household income and the likelihood of experiencing clinical depression?," and estimate the magnitude and direction of that relationship using a logistic regression model. In the "algorithmic modeling" culture, which aligns with much of ML-based research, the focus is on building a model that reliably maps input data to output data. In this culture, the parameters do not necessarily need to provide a faithful and interpretable description of patterns in the world; rather, the goal is to accurately predict output data when given a new sample of input data that is separate from the original data sample in which the model was trained. For example, researchers might ask, "given all the data available about a person in a particular dataset, how accurately can we predict whether that person will experience clinical depression?," and test a variety of models to find the one with the best predictive accuracy. Breiman’s proposition was that more scientific research should use "algorithmic modeling" \cite{Breiman_TwoCultures}. Several responses to Breiman have argued for the value of merging the "two cultures" or moving iteratively between them \cite{LUNDBERG_researcher_reasoning, Neufeld2021_TwoToOne, Shmueli2021_TwoToMulti, hofman_integrating_2021, baiocchi2021, ogburn2021}. 

Since non-ML ("data modeling") methods are currently the standard approach in many scientific disciplines, explaining the motivation for using ML methods will help readers better understand a study’s goals. Considering the differences and similarities between Breiman’s "two cultures" may be useful to researchers when motivating the use of ML methods. Additionally, many recent articles provide guidance on the value of ML methods for science \cite{hofman_integrating_2021, LUNDBERG_researcher_reasoning, athey_machine_2019, Molina_Garip, Grimmer_ml_for_social, Rashidi2021, Beam2018, MCCOY2022252, Tarca2007}.

\section*{Module 2: Computational reproducibility}

Computational reproducibility refers to the ability of an independent researcher to get the same results as reported in a paper or manuscript. It is an essential part of computational research~\cite{stodden_reproducing_2015}.

Computational reproducibility can help independent researchers evaluate the findings in a paper and verify if they hold up under scrutiny. The availability of reproducibility materials has led to several errors being spotted~\cite{herndon_does_2014,herzog_retraction_2022,berenbaum_retraction_2021,neunhoeffer_how_2019,hofman_expanding_2021,vandewiele_overly_2021}.
Conversely, if the code and data for reproducing all results in a study are not available, identifying the precise sources of errors in a study becomes hard~\cite{ioannidis_repeatability_2009,verstynen_overfitting_2023,haibe-kains_transparency_2020}. 

\textbf{Current computational reproducibility standards fall short. }
Some journals require authors to make their computational reproducibility materials available post-publication without requiring these materials at the time of publication. However, such measures can miss the mark. Stodden et al. \cite{stodden_empirical_2018} attempted to contact the authors of 204 papers published in the journal \textit{Science} to obtain reproducibility materials. Only 44\% of authors responded. Similarly, Gabelica et al.~\cite{gabelica_many_2022} studied papers published in 333 open-access journals indexed on BioMed Central in January 2019. Of the 1,792 papers that claimed they would share data upon request, 1,669 did not share the data. That is, they were unable to get the data for 93\% of the papers. This indicates the importance of requiring computational materials at the time of publication rather than at the authors' discretion later.

Vasilevsky et al. \cite{vasilevsky_reproducible_2017} studied the data-sharing policies at 318 biomedical journals. They found that almost a third of these journals had no data-sharing policies in place. Even the journals that did have data-sharing policies did not have clear guidelines for authors to comply with their policies.

ML methods research has also struggled to ensure computational reproducibility~\cite{henderson_deep_2018, musgrave_metric_2020}. Gundersen and Kjensmo \cite{gundersen_state_2018} systematically analyzed 400 papers that were published at leading conferences. In addition to code and dataset availability, they evaluated the documentation of methods in a paper's text, for instance, whether the experimental setup is described. They found that \textit{none} of the 400 papers satisfied all of their reproducibility criteria, and in general, papers only satisfied 20-30\% of the criteria.

Pineau et al. \cite{pineau_improving_2022} found that only around half of the papers submitted to NeurIPS 2018, a leading ML conference, contained the code and data needed to reproduce results. To improve reproducibility, they introduced a checklist that was used in NeurIPS 2019. In the checklist, including reproducibility materials was optional but recommended. Still, after the checklist was introduced, over 75\% of the papers included reproducibility materials along with the submissions. Similar checklists have become the standard at several ML conferences~\cite{noauthor_aaai_nodate,noauthor_neurips_nodate,noauthor_icml_nodate}. 

While ML methods research differs from ML-based science in its goals and practices, we can learn from these experiences to emphasize the importance of computational reproducibility in ML-based science.

\textbf{Ensuring computational reproducibility in ML-based science is challenging. }ML methods used in scientific research can be complex and often require several packages and dependencies. This makes computational reproducibility challenging~\cite{gundersen_machine_2022}. Liu and Salganik describe their experiences ensuring computational reproducibility while editing a special issue in Socius on the Fragile Families Challenge~\cite{liu_successes_2019}. The Fragile Families Challenge was a prediction competition where multiple participants tried to predict children's life outcomes on the same dataset~\cite{salganik2020measuring}. The special issue published papers based on a few of the resulting models. Liu and Salganik wanted to ensure that the code and data alongside every publication were verified to ensure that they produced the same results as those presented in the paper.
Despite spending 13 months working on achieving computational reproducibility and exchanging dozens of emails with the authors, they were unable to verify the computational reproducibility of all papers. Even preliminary steps, like installing the correct versions of each package, were non-trivial when a large number of packages were used in a study. They could eventually verify the computational reproducibility of 7 of the 12 papers. They published the rest with the code and data available at the time of publication.

\textbf{Interventions adopted by journals. }
Journals have adopted several measures to improve computational reproducibility~\cite{nature_reporting_nodate, science_science_nodate,noauthor_journal_nodate}. The Transparency and Openness Promotion (TOP) standards introduced by the Open Science Foundation~\cite{nosek_promoting_2015} have a few sections that focus on computational reproducibility in scientific research. They are divided into three levels: Level 1 requires stating if the computational reproducibility materials are available. If they are, authors should provide details about how to access them. Level 2 requires the materials to be available in a trusted repository at the time of publication. Level 3 requires the materials to be verified by the journal to ensure they generate the results reported in the paper prior to publication.

Several social science journals use the Data and Code Availability Standards for computational reproducibility~\cite{koren_miklos_data_2022}. Other journals have taken additional measures to verify if the code is correct. For instance, \textit{Nature Methods} conducts code reviews in addition to peer reviews for papers that provide computational artifacts~\cite{noauthor_reviewing_2015}.

In our checklist, we include the basic details that can enable independent researchers to verify the computational reproducibility of a result. 
Our checklist requires information about the dataset, code, computing environment, documentation for how to get the results in a study, and a reproduction script to automatically run the code and generate results. 
We acknowledge that computational reproducibility is hard and that some items in this module are more challenging compared to others. For instance, providing a reproduction script is not always possible. There may be other challenges. For instance, it is not always possible to release private datasets. We provide some options to address these challenges below, such as providing a synthetic imitation of the data when the real data cannot be publicly released.

    \textbf{2a) Dataset.} 
    Peng et al. \cite{peng_mitigating_2021} and Nosek et al. \cite{nosek_promoting_2015} highlight the importance of citing datasets with permanent links to clarify which version of a dataset is used in a study.
    If an original dataset is provided alongside a study, documentation for the dataset is also important. For instance, authors can include data dictionaries~\cite{noauthor_data_nodate} or datasheets~\cite{gebru_datasheets_2021}. Such documentation should report basic details about the properties and format of the data. 
    
    Some datasets could contain sensitive information and cannot be publicly released. To address this, authors have previously released synthetic datasets when working with sensitive data. For instance, Obermeyer et al. released a synthetic dataset alongside their study of racial bias in a healthcare algorithm~\cite{obermeyer_dissecting_2019}. They used the synthpop package in R to generate the synthetic data~\cite{nowok_synthpop_2016}.
    
    \textbf{2b) Code.} Similarly, it is important to report the exact version of code used for running the experiments and producing the results in a paper~\cite{sandve_ten_2013}. Authors can accomplish this by providing a DOI, commit tag (for instance, from code repositories such as GitHub, GitLab, or BitBucket), or other documentation to precisely identify the version of the code used to train and evaluate the model and produce the results reported in the paper. Note that using archiving systems that provide permanent identifiers for the code used, like Dataverse, is likely to aid long-term reproducibility~\cite{VINES201494, gibney_scientists_2013}.

    \textbf{2c) Computing environment.} Different computational experiments require different amounts of computing resources. To help readers understand the precise computing requirements for reproducing the study, authors should report details about the hardware (CPU, RAM, disk space), software (operating system, programming language, version number for each package used), and computing resources (time taken to generate the results) used to generate their results.
    Stodden \& Miguez \cite{stodden_best_2014} provide best practices to document computing infrastructure. 
    
    \textbf{2d) Documentation.} 
    Good documentation helps researchers unfamiliar with a project by walking them through the steps of setting up and running the code provided, starting from environment requirements and installation, to examples of usage and expected results~\cite{vilhuber_lars_template_2020, singers_awesome_nodate}.
    
    \textbf{2e) Reproduction script.} A script to produce all results reported in the paper using the reproducibility materials can significantly reduce the time it takes for an independent researcher to reproduce the results reported in a study.
    Reproduction scripts can download packages with the  version numbers needed to run the code, set the right dependencies, download and store datasets in the correct location, set up the computing environment, and run the code to produce the results reported in the paper. 
     
    Authors can implement such scripts in several ways, such as using a bash script~\cite{harbert_bash_2018} or using an online reproducibility platform such as CodeOcean~\cite{clyburne-sherin_computational_2019}. 
    Note that this is a high bar for computational reproducibility. In some cases, it might not be possible to provide a script that would allow an independent researcher to reproduce all results---for instance, if the analysis is run on an academic high-performance computing cluster, or if the dataset does not allow for programmatic download.

\section*{Module 3: Data quality} 
    
This module helps readers and referees understand and evaluate the quality of the data used in the study. Using poor-quality data or data that is not suitable for answering a research question can lead to results that are meaningless or misleading.

\textbf{3a) Data source(s).} Describing a study’s data source(s) allows readers to evaluate the data’s strengths and weaknesses and to judge whether the data are appropriate for the study’s goals. Most studies provide descriptions of their data source, but those descriptions sometimes lack important details \cite{Navarro2022_Completeness, Yusufe034568, Kim2016}. Additionally, the quality of reporting about ground-truth annotation methods in ML-based science varies widely \cite{Geiger2020}. To ensure a minimum level of information about data sources is reported, our checklist asks researchers to report when, where, and how data were collected, and how ground-truth annotations were performed on the dataset, if applicable.

\textbf{3b) Sampling frame.} A sampling frame is a list of people or units from which a sample is drawn. Due to practical limitations, the sampling frame in many studies does not include all members of the target population. It is important to describe a study's sampling frame so that readers understand the boundaries of the study's sample and how that sample relates to the target population (discussed further under checklist item \textit{3g: Dataset for evaluation is representative}). 

Some research papers do not provide a clear description of their sampling frame or eligibility criteria for inclusion in the sample \cite{Navarro2022_Completeness, Tooth2005, Porzsolt}. Furthermore, one review article found that when papers stated information about their sampling frame or eligibility criteria was available in a prior publication, the prior publication was not always accessible and the relevant information was often extremely difficult or impossible to find \cite{Porzsolt}.

\textbf{3c) Justification for why the dataset is useful for the modeling task.} Our checklist asks researchers to justify why the dataset is useful for the modeling task because the appropriateness of a data source will depend on the research question. For example, while biased or incomplete data is inappropriate for some research questions, such data can work well for other research questions as long as the researcher understands how these shortcomings impact the analysis and communicates the limitations~\cite{Grimmer_Roberts_Stewart_2022}. A broad claim like "this is the best dataset available on this topic" does not help readers understand the strengths and weaknesses of the data for the study's research question; researchers should be specific about \textit{why} the dataset is well-suited to the question.

Modern ML-based research often relies on repurposed data sources, which are sometimes termed "big data:" for example, social media data, digital trace data, or digital administrative records \cite{Salganik_BBB}. Salganik describes ten common characteristics of "big data" that result in differing strengths and weaknesses compared to traditional data sources \cite{Salganik_BBB}. Researchers who are using ML with repurposed data can use these ten characteristics as a guide when justifying why the data source is appropriate for their research goals and identifying shortcomings of the data.

\textbf{3d) Outcome variable.} Our checklist asks researchers to report how their outcome variable is defined. The outcome or target variable is the quantity that the model is used to predict, detect, classify, or estimate.

The outcome variable is typically an empirical proxy for an unobservable theoretical construct \cite{barocas-hardt-narayanan}. For example, researchers might pose a question about the construct "academic performance," and use grade point average as measured in school administrative data as the empirical proxy for this construct. The outcome variable is usually not a perfect match for the theoretical construct it represents. Thus, in order to allow readers to evaluate a paper’s claims, authors should describe precisely how their outcome variable is measured, and note any ways in which this outcome variable might not align with the associated construct. This is especially important because mismatches between variables and the constructs they are purported to represent can create fairness issues \cite{Jacobs_measurement}.

Our checklist also asks for descriptive statistics about the outcome variable. Reviews of prior literature have found that descriptive statistics are not always sufficiently reported \cite{crede_harms_2021, LarsonHall2015}. Descriptive statistics about the outcome variable help readers to understand the context being studied and to identify concerns related to rare or skewed values.

\textbf{3e) Number of samples in the dataset.} Reporting sample size is important because a study must have sufficient sample size to achieve its objectives. Some research objectives can be achieved with small to moderate-sized samples: for example, detecting large differences between groups. Other objectives generally require large samples: for example, studying rare events, studying heterogeneity, or detecting small differences \cite{Salganik_BBB}. 

Note that there can be downsides to large sample sizes. When the sampling frame is unrepresentative, increasing the sample size can shrink confidence intervals without decreasing bias in estimates, thus giving false confidence ~\cite{Bradley2021}. Furthermore, if a study exposes participants to any level of risk, larger samples may magnify harms \cite{Salganik_BBB}.

Scientific literature is generally consistent about reporting total sample size \cite{plonsky_2013, Tooth2005}. However, reporting on sample size for subgroups or sample size after attrition in longitudinal research is less consistent \cite{Tooth2005}. To ensure clear reporting, our checklist asks that in addition to the total sample size, researchers who are conducting a classification task report the number of samples in each class. This follows recent calls for more granular details about data, such as by Gebru et al. \cite{gebru_datasheets_2021}. We also ask researchers to distinguish between the number of individuals in the dataset and the number of rows in the dataset, in cases where an individual can appear in more than one row.

\textbf{3f) Missingness.} Missing data is highly prevalent in many research domains and can impact the results \cite{McKnight, Peugh, Salganik_FFC_appendix}. It is important for researchers to report the prevalence of missing data in their dataset and to specify how they handled missingness \cite{McKnight}. Missingness is particularly important to address carefully when it is non-random \cite{Mack}. Extensive literature across multiple fields has established that research articles frequently provide insufficient information about the presence and handling of missing data \cite{Nijman2022, Navarro2022_Completeness, Little2013, Nicholson2016, Sterner2011, Peugh, Tooth2005, Hussain2017}.  

Checklist item \textit{3f} focuses on reporting the prevalence of missing data. Reporting how missing data is handled is covered in item \textit{4c: Data transformations}.

\textbf{3g) Dataset for evaluation is representative.} This item asks researchers to justify that their sample is representative of the target population defined in Module 1. Representativeness is important for the study’s ability to generalize from the sample to the target population. Lack of representativeness in the sampling process is sometimes underreported; for example, one review of past literature found underreporting of information about selection bias \cite{Tooth2005}.  

Probability sampling is a common approach for achieving representativeness. However, probability sampling is not always necessary. Due to coverage errors and nonresponse in probability sampling, the differences between probability sampling and non-probability sampling are not always as large as they first appear \cite{Salganik_BBB}. For studies that use non-probability sampling, researchers may be able to make a reasonable argument for representativeness by comparing sample characteristics with population characteristics \cite{Wilkinson1999StatisticalMI}. Researchers can also use statistical methods to adjust for non-probability sampling (or to adjust for errors in probability sampling), such as post-stratification, sample matching, propensity score weighting, and calibration \cite{Salganik_BBB}. 

In some studies, the dataset will not be representative of the target population. Reasons a sample might fail to be representative of the target population depend on the type of data collected. For example, Salganik describes three types of representation error in survey data \cite{Salganik_BBB}, and Grimmer et al. describe four sources of bias in sample selection for text analysis \cite{Grimmer_Roberts_Stewart_2022}. A non-representative sample is okay for certain research goals. For example, Salganik argues that research that aims to make out-of-sample generalizations generally requires representative data, but research that aims to make within-sample comparisons can be well-suited to non-representative data \cite{Salganik_BBB}. Concerns about non-representative\-ness should be noted under checklist item \textit{8a: Evidence of external validity.}

    \section*{Module 4: Data preprocessing} 

Preprocessing is the series of steps taken to convert the dataset from its rawest available form into the final form used in the modeling process. This includes data selection (i.e., selecting a set of samples from the dataset to be included in the modeling process) as well as other transformations of the data, such as imputing missing data and normalizing feature values.

Our checklist focuses on two broad components of preprocessing: first, the subset of data to consider (i.e., which rows of a dataset are considered), and second, the transformations that are subsequently applied to the data (i.e., how entries of a dataset might be altered). Each of these components have implications on the scope and validity of resulting scientific claims, and are essential for ensuring the reproducibility of the results. As discussed in Module 3, preprocessing methods are often not specified in papers.

\textbf{4a) Excluded data and rationale.}
Researchers might exclude some samples from the dataset—for instance, to remove outliers or to only focus on certain subsets. Thus, the resulting scientific claims should be made in relation to the particular subset of a dataset that is ultimately used. This type of preprocessing is closely related to our discussion on data quality (Module 3) and generalizability (Module 8). This item underscores the importance of reporting the specific subset of the dataset used, in addition to details about the overall data.

Hofman et al. \cite{hofman2017prediction} note how the choice of the subset can significantly affect the performance of the resulting model. A specific example they use is the prediction of the reach (``cascade size'') of a social media post as a function of the poster’s past success. They show that the threshold of popularity used to determine the subset of posts considered plays a large role in influencing the predictability of cascade size. The scientific claim---regarding the predictability of the success of a post---depends on what data is included and excluded. We ask authors to justify why the particular subset of data used was chosen.

\textbf{4b) How impossible or corrupt samples are dealt with.}
A dataset may contain erroneous or undesirable data points. Data may be impossible (e.g., a person whose height is recorded as 10 feet) or corrupted (e.g., a survey response filled out by a bot). Different techniques have been developed to detect such cases \cite{chu2016data, buchanan2018methods}. Attempting to filter such data can be important to ensure the dataset represents the intended population of the study. Data detected as impossible or corrupt may be removed or transformed---our checklist asks authors to report such steps in items \textit{4a} and \textit{4c} respectively.

\textbf{4c) Data transformations.}
Once the set of data points to be used is decided upon, researchers could transform the data in various ways: for example, by normalizing or augmenting data, or imputing missing data. Both imputing missing data and under/over-sampling from a subpopulation---if done improperly---can harm the validity of model performance in relation to the scientific claim made. For example, mean imputation and oversampling must be done separately on each fold of a dataset. Failing to do so can result in overoptimistic results~\cite{vandewiele2021overly}. 

Specifying preprocessing steps is also important for ensuring the reproducibility of the results. Choices in preprocessing technique can significantly affect the properties of the resulting model, including accuracy and interpretability \cite{shadbahr2022classification, gryska2022deep}.

\section*{Module 5: Modeling}

Researchers make several choices in creating an ML model. The exact specification of the model is important to consider with respect to the particular scientific claim being made. In addition, due to the large number of choices involved in creating an ML model, it is important to report exact details of how an ML model is created---otherwise, reproducibility by independent researchers could be hindered. Raff \cite{raff_step_2019} attempted to reproduce ML results from 255 papers using only the paper's text (i.e., without using the code accompanying a paper), and found that 93 could not be reproduced.

Modeling choices are also closely related to choices involving evaluation. Indeed, model selection---the process of choosing the model(s) whose results are reported in a study---often depends on the evaluation setup. An improper approach to evaluation or model selection can result in exaggerated performance estimates.

Major machine learning conferences, such as NeurIPS and ICML, incorporate checklists that ask authors to verify that they have included training details within their paper, similar to the specific items we propose below~\cite{noauthor_icml_nodate,noauthor_neurips_nodate}. Similarly, Mitchell et al. introduced model cards to document details about the modeling and evaluation process, with a primary focus on natural language processing and computer vision~\cite{mitchell2019model}.

A note on terminology: there is no agreement on the use of different terms related to the modeling process. For the purpose of the discussion in this module, we make one conceptual distinction. Items \textit{5a} and \textit{5b} focuses on aspects of the model that are specified prior to training. This includes, for example, the type of model and the loss function. Meanwhile, items \textit{5c, 5d, 5e}, and \textit{5f} focus on fitted (trained) models.

\textbf{5a) Model description.} Specific details about all models trained are essential for ensuring the reproducibility of a paper. This includes specifying the input and output of a model, the type of model (e.g., Random Forest, Neural Network), and the loss function and algorithm used to train the model.

\textbf{5b) Justification for the choice of model types implemented.} With many different possible types of machine learning models to choose from, the choice of types to consider can be dependent on the intended scientific claim of a paper. For example, if a scientific claim aims to establish how predictive a set of features can be (e.g., \cite{salganik2020measuring, kleinberg2017theory}), then it may be appropriate to consider a wide range of types---the most important criteria is the resulting accuracy of the learned models. If the scientific claim aims to establish the potential usability of a model in practical settings, there may be additional desiderata. For example, a model that is used in high-stakes settings like clinical decision-making may need to be interpretable or explainable \cite{roscher2020explainable}, so it may be more appropriate to choose a model type that is likelier to have these properties. 

It is not always clear what the best type of model to use is for a given scientific question. For example, there is an active debate on how or if interpretability or explainability should be implemented in the context of different applications (see e.g., \cite{rudin2019stop}). Here, we ask authors to provide reasons for why the set of model types they implemented is appropriate.

The focus of this item is on the types of models considered, but some claims may depend on the choice of a specific learned model (and its performance). The following items address the process of evaluating, selecting, and comparing specific learned models.

\textbf{5c) Model evaluation method.} We ask authors to report details about the model evaluation procedure. We include evaluation in this `modeling' module since evaluation is often used as part of the model selection process (see item \textit{5d}). That is, it is common to consider many models and select the best-performing one based on the evaluation setup.

Evaluation of ML models must be done on test data separate from training data. To ensure reproducibility and verify validity, it is necessary to report how data was split and used. We ask authors to report whether models are evaluated using cross-validation, a holdout test set, or an external validation test set. We also ask for the sample size within each split of the data, including the number of samples of each class for classification tasks.

\textbf{5d) Model selection method.} There are many possible learned models that can arise depending on specific choices in the modeling process. Even holding the type of model fixed, differences can arise due to the specific choice of hyperparameters, for example. Researchers should report how the final model(s) reported in a paper are selected. 

A common goal of model selection is to select the model with the best performance on a hold-out set, but improper model selection can result in misleading performance~\cite{raschka_model_2020, cawley_over-fitting_2010}. 

Testing multiple models on the holdout set and choosing the one with the best performance on the holdout set can result in an over-optimistic estimate of performance. Neunhoeffer and Sternberg \cite{neunhoeffer_how_2019} consider this type of error in the context of political science, showing how a study on civil war prediction failed to reproduce because of improper model selection.

Model selection is not limited to choosing the model with the highest accuracy. Multiple models can achieve the same accuracy as one another yet make different predictions for the same individuals and possess different characteristics regarding issues like fairness and interpretability. This phenomenon is known as predictive multiplicity~\cite{marx_predictive_2020,watson-daniels_predictive_2023}. Black et al. provide guidance on opportunities and concerns that arise from such issues~\cite{Black2022}. 

\textbf{5e) Hyperparameter selection.} Training procedures are often dependent on the choice of model hyperparameters, such as regularization weight, the number of training epochs, or the learning rate for a model. The choice of hyperparameters is part of the model selection process. 

In the context of natural language processing research, Dodge et al. \cite{dodge2019show} show that details about hyperparameter search impact the resulting accuracy. More resources devoted to searching over hyperparameters can improve performance significantly. Islam et al. \cite{islam2017reproducibility} show that due to variance in performance depending on hyperparameters, specific choices of hyperparameters can be misleading when interpreting results. One method may perform better than another with one set of hyperparameters while performing worse given a different set of hyperparameters. Incorrect estimates of model performance due to hyperparameter optimization are especially concerning when comparing different models~\cite{cooper_hyperparameter_2021,sivaprasad_optimizer_2020,dahl_benchmarking_2023}. 

Finally, note that the degree to which hyperparameter optimization affects results can vary depending on the model type \cite{probst2019tunability}. This may be a reason to prefer one model type over another.

\textbf{5f) Appropriate baselines.} It can be important to compare the performance of a model to baselines, especially when the scientific claim argues that a particular ML method can outperform existing approaches on a task. To clearly establish the performance of an ML model, it is necessary to detail how baseline models were trained and how the baseline methods are optimized. For example, if the baselines were not chosen using the same model selection methods, this could result in the baselines being ``weak.'' As a result, the apparent benefits of the new method may be misleading. Sculley et al.~\cite{sculley_winners_2018} and Lin \cite{lin_neural_2019} detail examples in which machine learning models were compared to weak baselines. 

Lones \cite{lones_how_2023} discusses how to compare models more broadly. See also item \textit{7c: Appropriate statistical tests} for a discussion on the use of statistical testing to compare models.

\section*{Module 6: Data leakage} 

Leakage is a spurious relationship between the features and the target variable that arises as an artifact of the data collection, sampling, pre-processing, or modeling steps. For example, normalizing features in the training and test data together leads to leakage since information about the test data features is included in the training data~\cite{kaufman_leakage_2012}.  

Data leakage is a common error in the use of ML methods. Epic, a U.S. healthcare technology company, released a sepsis prediction model in hospitals nationwide. However, one of the features used in the model was whether a patient had been prescribed antibiotics. This is an error because antibiotics would typically be prescribed after the diagnosis of sepsis, so they act as a proxy for the outcome variable. Consequently, the model's performance was inflated due to having access to information it would not have in a real-world scenario~\cite{ross_epics_2021}. 

Leakage has caused widespread reproducibility errors in ML-based science. In a survey of leakage across ML-based science, Kapoor and Narayanan found that leakage affects hundreds of papers across 17 fields  ~\cite{kapoor_leakage_2022}. 

We ask authors to justify that their study does not suffer from major sources of leakage, which we elaborate on in this section. Kapoor and Narayanan~\cite{kapoor_leakage_2022} offer model info sheets to help detect and prevent leakage before publication. Our checklist focuses on the three main types of leakage in ML-based science found in their survey.

\textbf{6a) Train-test separation is maintained.} When information from the test set is used during the training process, it leads to over-optimistic performance results due to data leakage. 

Not using a held-out test set is a textbook error in ML~\cite{kuhn_applied_2013}. Still, it is widespread. For example, Poldrack et al. \cite{poldrack_establishment_2020} find that of the 100 neuropsychiatry studies that claimed to predict patient outcomes, 45 only reported in-sample statistical fit as evidence for predictive accuracy.

There can be other, more subtle variations of this error. For example, if the train-test split occurs after any of the other pre-processing or modeling steps (Modules 4 and 5), this also results in leakage. Vandewiele et al. \cite{vandewiele_overly_2021} found overoptimistic results in 21 papers that claimed to predict the risk of pre-term births. These papers suffered from the same error: oversampling data before partitioning it into training and test sets. This resulted in the test set becoming artificially similar to the training set, and led to exaggerated performance claims across the literature on pre-term risk prediction.

\textbf{6b) Dependencies or duplicates between datasets.} In some cases, samples in the dataset might have dependencies. For example, a clinical dataset might have many samples from the same patient. Oner et al. \cite{oner_training_2020} find that when image data from the same patient is present in training and test sets, it leads to overoptimistic results. Similarly, for time-series forecasting models, randomly splitting a time-series dataset into training and test sets is likely to lead to overoptimism, because the training data has information ``from the future''~\cite{malik_hierarchy_2020}. 

In such cases, the train-test split or cross-validation (CV) split should take these dependencies into account---for instance, by including all samples from each patient in the same train-test split or CV fold. 
There are several ways to avoid such dependencies. Bergmeir \& Benítez \cite{bergmeir_use_2012} find that blocked CV for time series evaluation deals with temporal autocorrelation. Hammerla \& Plotz \cite{hammerla_lets_2015} demonstrate how ``neighborhood bias'' can affect data recordings close in time. They introduce "meta-segmented CV" to deal with such dependencies. Roberts et al. \cite{roberts_cross-validation_2017} describe block CV strategies for a number of structures with dependencies, including temporal, spatial, and hierarchical dependencies.

Duplicates in the datasets can also spread across training and test sets if the dataset is split randomly. This should be avoided, as it leaks information across the train-test split. Roberts et al. \cite{roberts_common_2021} outline this error with Frankenstein datasets: datasets that combine multiple other sources of data can end up using the same data twice---for instance, if two datasets rely on the same underlying data source are combined into a larger dataset.

\textbf{6c) Feature legitimacy.} If any of the features used in a model is a proxy for the outcome, this can result in leakage. Filho et al. \cite{chiavegatto_filho_data_2021} found that a prominent paper on hypertension prediction \cite{ye_prediction_2018} suffered from data leakage due to illegitimate features. The model included the use of anti-hypertensive drugs as a feature to predict hypertension. But this feature is not available when predicting if a new patient suffers from hypertension in a clinical setting, so it artificially inflates the performance of the ML model. Similarly, Epic's sepsis prediction tool used ``antibiotics'' as a feature to predict if someone would get sepsis ~\cite{ross_epics_2021}.

This type of leakage is more likely when there are a large number of features, due to the increased likelihood of including one or more illegitimate features. The sheer volume of features can make it challenging to scrutinize each one for potential leakage.

\section*{Module 7: Metrics and uncertainty quantification}
The performance of ML models is key to the scientific claims of interest. Since authors can make many possible choices when choosing performance metrics, it is important to reason why the metrics used are appropriate for the task~\cite{leist_mapping_2022,forde_model_2021}. Additionally, communicating and reasoning about uncertainty is important, but uncertainty is currently under-reported~\cite{bhatt_uncertainty_2021,cooper_is_2023,qian_are_2021,young_model_2018}. For example, Simmonds et al.~\cite{simmonds_how_2022} find that studies often do not report the various kinds of uncertainty in the modeling process.

We ask authors to report their performance metrics and uncertainty estimates for those metrics in enough detail to enable a judgment about whether they made valid choices for evaluating the model's performance.

The checklist requires authors to detail how they evaluate the performance of their model on its own and in relation to baselines. In addition, in this section, we ask authors to reason about their measurement of model performance and uncertainty in relation to their scientific claim.

\textbf{7a) Performance metrics used.} There are many possible ways to measure the performance of an ML model (see Table 4 in \cite{leist_mapping_2022} for an overview). Certain metrics can be misleading or inappropriate. For example, accuracy might not be suitable to measure the performance of an ML model in the presence of heavy class imbalance: when most data points have positive labels, it is easy to obtain high accuracy simply by predicting positive for all cases \cite{monard2002learning}. Other common metrics, like AUC, also have limitations \cite{lobo2008auc}. 

The proper choice of metric is often influenced by the particular application being studied. In some domains, certain errors may be more costly than others. For example, false positives are more costly than false negatives in email spam detection~\cite{bhowmick2018mail}. We ask authors to justify the use of a particular performance metric in relation to the scientific claim.

\textbf{7b) Uncertainty estimates.} Uncertainty in the performance of ML models can arise in many ways. For example, it could arise from randomness in the training data, evaluation data, or in the training process itself. Uncertainty is important to capture when evaluating the strength of a scientific claim. Because a dataset represents only a finite sample from a population, there is uncertainty in this sampling. Since it is generally infeasible to obtain several independent samples, methods such as bootstrapping (evaluating model performance by re-sampling from the same dataset) are often used to generate confidence intervals~\cite{raschka_model_2020}.

In a systematic review of uncertainty quantification across seven scientific fields, Simmonds et al.~\cite{simmonds2022model} found that scientific fields differ in what kinds of uncertainty they report. They propose best practices and a checklist to help researchers account for uncertainty.

\textbf{7c) Appropriate statistical tests.} Comparing the performance of different models can be key to some scientific claims. For example, a scientific claim may argue that one ML method outperforms others. Statistical testing is one tool to evaluate such differences in performance. Raschka \cite{raschka_model_2020} gives an overview of statistical tests for ML models. Still, in some or even most cases, an appropriate and valid statistical test may not be known; this can be an area for further development.
The reliance on statistical significance testing has also led to misinterpretations and false conclusions. As a result, reporting uncertainty is better than performing statistical tests alone~\cite{amrhein2019scientists}. Still, if a statistical test must be performed, it should be appropriate for the comparison.

\section*{Module 8: Generalizability and limitations} 
    
 External validity (or "generalizability") refers to the extent to which the findings from a study’s sample apply to the target population, as well as the extent to which the findings apply to other populations, outcomes, and contexts ~\cite{Shadish2001, EGAMI2022}.

ML-based science faces a number of threats to external validity \cite{hullman_worst_2022, raji_fallacy_2022}. Since studies that use ML methods are often unaccompanied by external (i.e., out-of-distribution) validation \cite{liao_are_2021}, it is important to reason about these threats. Additionally, authors are best positioned to identify the boundaries of applicability of their claims in order to prevent misunderstandings about the claims made in their study.

\textbf{External validity of three types of claims in ML-based science.} We distinguish among three types of external validity, corresponding to three types of claims made in ML-based science.

\textit{External validity of claims about observed patterns.} In some studies, researchers use ML methods to make a claim about the presence of a pattern in the world. For example, Mathur et al. use structural topic modeling to study manipulative tactics in a corpus of U.S. political campaign emails from the 2020 election cycle. One of their findings is that the median active sender of campaign emails uses "sensationalist clickbait" in 37\% of those emails \cite{Mathur2023}. A question about the external validity of this claim might be: is the frequency of sensationalist clickbait in campaign emails similar in other U.S. election cycles? 

\textit{External validity of claims about learned models.} A "model" is a function that an algorithm learns when the algorithm is applied to data \cite{Brownlee}. In some studies, researchers train an ML model and make a claim about the model’s performance when deployed in real-world settings. For example, Dugas et al. build a model to forecast influenza outbreaks at the city level using Google Flu Trends and other readily accessible data, with the goal that the model can be deployed by medical centers to provide warning of upcoming outbreaks. They find that "the model, on the average, predicts weekly influenza cases during 7 out-of-sample outbreaks within 7 cases for 83\% of estimates" (p. 1) \cite{Dugas2013}. A question about the external validity of their claim, which they note in their discussion, is: does the model achieve similar performance in other years? External validity of claims about learned models is more widely known as "domain generalization" or "robustness to distribution shift" and is a well-studied phenomenon in ML \cite{Wiles, Koh}. 

\textit{External validity of claims about learning algorithms.} An "algorithm" is a procedure for learning from data \cite{Brownlee}. In some studies, researchers make a claim about the usefulness of ML algorithms in a particular context. For example, Bansak et al. develop an ML algorithm to assign refugees to resettlement sites by "leverag[ing] synergies between refugee characteristics and resettlement sites." They test the algorithm in retrospective data from the U.S. and Switzerland, and find that their "approach led to gains of roughly 40 to 70\%, on average, in refugees’ employment outcomes relative to current assignment practices" (p. 1). They claim that governments can use this algorithmic assignment approach to improve resettlement outcomes for refugees. However, unlike the influenza forecasting example above, they do not claim that the specific model they trained can be deployed directly into use \cite{Bansak2018}. A question about external validity of their claim might be: does algorithmic assignment of refugee resettlement locations work similarly well for other time periods? 

\textbf{Reporting on external validity falls short in past literature.} Reviews of past literature have found that scientific papers sometimes lack discussion about external validity \cite{Tooth2005} or lack information about sample demographics that would help readers draw their own conclusions about external validity \cite{Bozkurt2020}. Furthermore, many ML-based studies make claims about the ability of a model to deploy in real-world settings, but do not report whether their evaluation sample matches the population in the real-world setting \cite{Bozkurt2020}, and do not conduct external validation \cite{Bozkurt2020, Navarro2023}.

\textbf{8a) Evidence of external validity.} In this checklist item, we ask researchers to discuss the ability to generalize their claims from the sample to the target population and to other populations, outcomes, or contexts. Researchers can use a mix of quantitative and theoretical approaches to make arguments regarding their findings' external validity. They can report quantitative evidence by testing their claims in out-of-distribution data. They can make theoretical arguments about their expectations of external validity by referring to prior literature and reasoning about the level of similarity between contexts ~\cite{Simons2017}. 

Several threats to the external validity of ML models have been documented in past literature. Finlayson et al. outline external validity failures due to dataset shifts in clinical settings \cite{Finlayson2021}. Hullman et al. discuss the threats to external validity that arise in different phases of an ML research project \cite{hullman_worst_2022}. Liao et al. outline a taxonomy of evaluation failures in ML, including failures in external validity \cite{liao_are_2021}. Geirhos et al. discuss the phenomenon of shortcut learning in ML models, a phenomenon where models rely on shortcuts (such as the background color in an image) instead of detecting patterns that actually relate to the phenomena of interest \cite{Geirhos2020}.

Researchers who make claims about learned models should be aware that even if their model currently generalizes to their target population, that performance may degrade over time due to temporal distribution shift (or "temporal drift"). Causes of temporal drift include changes in technology, changes in population and setting, and changes in behavior \cite{Finlayson2021}. Concerns about the risk of drift should be communicated in the paper, when applicable.

\textbf{8b) Contexts in which the authors do not expect the study’s findings to hold.} Clarifying the circumstances under which scientific conclusions or model performance are not expected to hold helps to set clear expectations and avoid unjustified hype. Raji et al. find that flaws in ML models deployed in real-world settings stem in part from a lack of focus on identifying when models are not expected to work \cite{raji_fallacy_2022}. Simons et al. argue that making an explicit "constraints on generality" statement that identifies boundaries of the circumstances where findings are expected to hold has several benefits, including helping to ensure that a study's conclusions accurately reflect its evidence, increasing the likelihood of successful replication, and inspiring follow-up studies that test the findings in new populations~\cite{Simons2017}.

\section*{Conclusion}

ML methods present an exciting advance for scientific research. Done right, they can allow researchers to analyze complex data and work with new modalities such as images and video. Yet, recent failures of ML-based science reveal the urgent necessity of improving standards of transparency across fields that use ML methods.
Our paper provides a cross-disciplinary bar for reporting standards in ML-based science.

\textbf{Supplementary materials.} All appendices are available at \href{https://reforms.cs.princeton.edu/}{https://reforms.cs.princeton.edu/}.

\textbf{Acknowledgments.} We thank A. Feder Cooper for their feedback on the paper.

\vspace{-0.3cm}

\bibliographystyle{unsrt}
\bibliography{references, references-1, references-2, references-kenny,references-3}

\end{document}